\documentclass{article}
\usepackage{spconf,amsmath,graphicx}
\usepackage[table,x11names,dvipsnames]{xcolor}
\usepackage{pgfplots}

\usepackage{tabularx}

\pgfplotsset{compat=1.18}

\title{NO PITCH LEFT BEHIND:
ADDRESSING GENDER UNBALANCE \\ IN AUTOMATIC SPEECH RECOGNITION THROUGH PITCH MANIPULATION}

\name{Dennis Fucci\textsuperscript{1,2}, Marco Gaido\textsuperscript{1}, Matteo Negri\textsuperscript{1}, Mauro Cettolo\textsuperscript{1}, Luisa Bentivogli\textsuperscript{1}}

\address{\textsuperscript{1}Fondazione Bruno Kessler, Italy\\
\textsuperscript{2}University of Trento, Italy\\
\texttt{\{dfucci, mgaido, negri, cettolo, bentivo\}@fbk.eu}}
\copyrightnotice{979-8-3503-0689-7/23/\$31.00~\copyright2023 European Union}
\begin{document}
\maketitle
\begin{abstract}
Automatic speech recognition (ASR) systems are known to be sensitive to the sociolinguistic variability of speech data, in which gender plays a crucial role. This can result in disparities in recognition accuracy between male and female speakers, primarily due to the under-representation of the latter group in the training data. While in the context of hybrid ASR models several solutions have been proposed, the gender bias issue has not been explicitly addressed in end-to-end neural architectures. To fill this gap, we propose a data augmentation technique that manipulates the fundamental frequency ($f0$) and formants. This technique reduces the data unbalance among genders by simulating voices of the under-represented female speakers and increases the variability within each gender group. Experiments on spontaneous English speech show that our technique yields a relative WER improvement up to 9.87\% for utterances by female speakers, with larger gains for the least-represented $f0$ ranges.
\end{abstract}
\begin{keywords}
Gender Bias, Data Augmentation, ASR. 
\end{keywords}
\section{INTRODUCTION}
\label{sec:intro}

As society increasingly relies on automated systems for processing
textual and audio data,
ensuring unbiased and inclusive outputs has become a critical concern.
One area of language technology that has garnered significant attention in this regard is 
automatic speech recognition
(ASR),
where inherent biases and errors persist 
despite remarkable advancements in the task.
Sociolinguistic variability, determined by demographic factors such as age, gender, ethnicity, and dialect \cite{Trudgill2001}, 
has historically 
posed challenges
for ASR models, which are biased toward more frequent patterns in their training data
\cite{TatmanK17, Moro-VelazquezC19, feng2021quantifying}.
Among demographic factors, gender\footnote{
In this paper, we use ``gender'' as a simplified approximation for sex-related differences within a binary framework (female and male) to address the anatomical factors contributing to phonetic variability. We acknowledge that phonetic variability among gender groups can also arise from social factors that do not neatly fit into this binary framework. However, in this study, we do not investigate these social aspects.
}
is one of the most prominent, as
female voices are often disadvantaged in favor of better recognition performance on male voices.
This phenomenon has been observed in various instances such as YouTube's automatic captions \cite{tatman-2017-gender},\footnote{However, gender bias was not observed in a second study \cite{TatmanK17}.} hybrid HMM-DNN architectures applied to French broadcast data \cite{Garnerin19}, and DNN systems used for English read speech \cite{Garnerin21}.
Although there have been sporadic reports of higher recognition accuracy for female voices \cite{Adda-DeckerL05}, these outliers
were
attributed to
the consistent adherence of female speakers to standard pronunciations, as opposed to the higher proportion of disfluencies
among male speakers within specific datasets \cite{Adda-DeckerL05}.

The key factor contributing to this discrepancy between female and male voices is the under-representation of the former in the training data \cite{Garnerin19}.
However, efforts
to address the gender-related performance gap have not explicitly focused on tackling the data imbalance; rather,
specific training strategies have been explored.
In the case of HMM and hybrid ASR systems, a common approach is the creation of gender-dependent models \cite{konig, Abdulla1988ImprovingSR, sallam}, which are designed to handle a specific gender without requiring to generalize across them, thereby reducing the input space
and potentially improving performance.
Additionally, 
speaker normalization  \cite{Giuliani, krishna18_sltu}
and speaker adaptive \cite{Bell, survey} techniques have been employed to
address
speaker variability without explicitly considering gender.
Despite these efforts, the problem of gender bias in ASR systems 
persists \cite{Garnerin19, Garnerin21}
and the specific 
issue of data imbalance 
has not been addressed with the advent of DNNs.

Along this direction, 
we propose a data augmentation technique specifically tailored to enhance the
robustness of ASR models to the less-represented female voices.
Inspired by other works on pitch\footnote{Throughout the paper, we employ the term ``pitch'' to denote the perception of vocal height,
with no implicit reference to any specific acoustic parameter (e.g., the fundamental frequency).} perturbation strategies \cite{jaitly, yeung-f0based, kathania-lpc1}, our approach is based on
the manipulation of the fundamental frequency ($f0$) and the formants of a speech segment,
which are critical acoustic parameters distinguishing male and female speech \cite{Coleman, Hillenbrand}.
The objective of these perturbations is to shift the pitch of a segment spoken by a speaker of one gender towards characteristics
typical either of the same gender or of the opposite gender, thereby increasing the variability within gender groups, in particular females, and decreasing the representation gap between the two groups. 
Experiments conducted on English spontaneous speech 
demonstrate
the effectiveness of our method in improving the performance of both male and female gender groups.
Notably, the method outperforms
a baseline system trained on the original non-manipulated data,
yielding relative improvements for both genders 
(up to 9.87\% for the under-represented female group).
Lastly, a focused
analysis of the outputs unveils that the highest gains are achieved
for
$f0$
values that are less represented in the training data,
emphasizing the potential of our method to build
more inclusive and fair systems.

\section{RELATED WORKS}

Data augmentation refers to increasing
the amount and variability of the training data by generating additional examples. 
By doing so, data augmentation techniques aim at increasing
the generalization 
capability
of the trained models so that they are more robust to a wider range of inference-time inputs and conditions. This is particularly important for neural systems,
which require large training datasets and risk overfitting \cite{Ying_2019}.

In ASR, various data augmentation techniques have been explored, in particular after the advent of end-to-end 
neural systems. They
mostly involve altering the original audio or its feature representation to generate instances with distinct acoustic characteristics.
In particular, we can isolate four approaches: \textit{noise augmentation}, \textit{time-scale modification}, \textit{time and frequency masking}, and \textit{pitch perturbation}.
\textit{Noise augmentation} consists in adding various types of noise to the clean speech signals
in order to
enhance the robustness of acoustic models to background noise present in real-world environments \cite{hannun-noise, ragni-noise}. As such, they are unrelated to the robustness of a system with respect to different speaker voices, which is the focus of our work. 
\textit{Time-scale modification} involves altering the speech tempo -- possibly of only a part of the original utterance, e.g. to simulate prolongation of vowels \cite{nagano-tempo}  -- while preserving the spectral envelope, and therefore the timbre, of the original signal \cite{ko, disordered}. For this reason, they do not address the variability of the speaker pitch. 
Lastly, the most popular method is based on \textit{time and frequency masking}, which consists in randomly masking time segments and frequency bins in the spectrogram, so as to enhance the model ability to handle noise and irrelevant temporal and spectral variations in the audio signal \cite{specaugment}. Specifically, thanks to its effectiveness and efficiency, SpecAugment \cite{specaugment} has become a standard component in training recipes of the most popular toolkits used to develop state-of-the-art ASR systems, like NeMo,\footnote{https://github.com/NVIDIA/NeMo.} ESPnet,\footnote{https://github.com/espnet/espnet.} Fairseq,\footnote{https://github.com/facebookresearch/fairseq.} and SpeechBrain.\footnote{https://github.com/speechbrain/speechbrain.} 
However, SpecAugment is not specifically designed to consistently obtain the vocal variability required to train gender-fair models.

More in line with our objective of increasing gender variability,
\textit{pitch perturbation}
modifies the spectral envelope of the speech signal, 
introducing variations in the vocal characteristics of the speaker to improve
the generalization capability of the model
across speakers and natural pitch variations.
This is typically done through the so-called ``vocal tract length perturbation'' (VTLP) scheme, 
which applies a warping function to the frequency axis of the spectrum 
to alter the pitch
while keeping the audio duration unchanged \cite{jaitly, ragni, kanda-elastic, disordered}.
Various warping equations can be used for pitch perturbation. 
The piecewise linear rule and the bilinear rule are commonly employed methods \cite{improved-vtlp},
but non-linear functions based on the tonotopic distance between formants and
$f0$
have also been proposed \cite{yeung-f0based}.
Often, the warping is also
performed by modifying the original linear predictive coding (LPC) spectrum \cite{kathania-lpc1, kathania-lpc2, kathania-lpc3, dua-lpc-tts}. 
In some cases, a random perturbation may be applied to each pole of the LPC synthesis filter \cite{lpc-augment} or to each segment of the spectrum \cite{vishwanath-lpc}.
Regardless of the type of warping function and the features to which they are applied,
pitch manipulation techniques have 
demonstrated effectiveness in addressing specific challenges in ASR, such as
low-resource languages \cite{ragni, kanda-elastic},
disordered speech \cite{disordered},
and child speech \cite{kathania-lpc1, kathania-lpc2, kathania-lpc3, vishwanath-lpc, lpc-augment, dua-lpc-tts, yeung-f0based}.
However, it has not been specifically employed to address the issue of gender bias, as we do here, and it typically involves a warping of all the frequencies, while here we only focus on the acoustic parameters that mostly differentiate males and females ($f0$ and formants).

\section{PITCH MANIPULATION}
\label{sec:methodology}

It is widely recognized that male and female voices exhibit distinct acoustic characteristics attributed to differences in vocal tract length and anatomy.
Two crucial parameters for distinguishing between male and female voices are $f0$ 
and formants \cite{Coleman, Hillenbrand}.
Therefore, 
to fulfill our objective of enhancing the representation of gender groups in the training data,
and specifically the under-represented female voices, 
we devise a data augmentation approach 
that revolves around perturbing 
these two acoustic parameters
in the audio segments of the training data.
Specifically, we propose two policies that aim at increasing the diversity of training data of each gender group and rebalancing the 
female/male voice ratio.
In the following, we first describe our two policies (\S\ref{sec:policies}), then we delve deep into the technical description of how the audio 
manipulations
are performed (\S\ref{sec:pitch_format_shifting}).

\subsection{Policies}
\label{sec:policies}

We define two policies, namely \textit{Opposite} and \textit{Random}, to determine whether an audio sample should be altered and how.
The policies are applied independently over the training epochs to each sample, so as to maximize the variability of the training data and avoid additional storage requirements (as the manipulated audio segments are computed on-the-fly and never saved). 
Both policies rely on knowing the speaker's gender for each utterance.
In our experiments (see \S\ref{sec:data}), we use
training data that already includes information about the speakers' gender.
However, when this information is not available, it can be inferred based on a simple spectral analysis
(e.g., by considering the averaged $f0$ value and comparing it to the typical $f0$ ranges for male and female speakers).

\noindent\textbf{Opposite.} 
This policy aims at rebalancing the amount of 
female/male
training samples
by shifting both the 
$f0$
and formants to values that are typical of the opposite gender. Specifically, we control the 
female/male
data ratio by means of two probabilities, $p_{f\rightarrow{}m}$ and $p_{m\rightarrow{}f}$, which respectively determine the probability of manipulating a segment uttered by a female voice (toward a male voice) and the probability of manipulating a segment uttered by a male voice (toward a female voice).
As such, by choosing $p_{f\rightarrow{}m} < p_{m\rightarrow{}f}$, we can increase the proportion
of female training samples.

\noindent\textbf{Random.}
Under
this policy, the vocal traits of the speaker are manipulated to obtain either a voice typical of the opposite gender or a voice typical of the same gender group. This decision is taken with a 50\% probability for each option. 
In the first case, both 
$f0$
and formant shifting are performed, like in the \textit{Opposite} policy.
In the second case, 
where
manipulation takes place
within the same gender group,
we only shift the $f0$
within each gender-specific frequency range without altering the formants, as preserving the formant characteristics contributes to maintaining the distinct vocal traits of each gender.
The selection of the target gender for augmentation is done randomly and independently from the original speaker gender. 
This approach aims to reduce the skew in the female/male training data ratio, 
promoting gender balance and simultaneously increasing
the overall variability of the data within each gender group.
However, since only the portion of the training data that 
undergoes manipulation
becomes balanced, while the unmodified part remains biased,
this policy does not achieve perfect balancing of the training data between gender groups. 
The extent of data that undergoes manipulation is determined by the single hyperparameter $p_r$, i.e.
the probability of applying augmentation to a sample: the higher $p_r$ is, the more balanced the training data is.

\subsection{$f0$ and Formants Shifting}
\label{sec:pitch_format_shifting}

Once the chosen policy has defined how the input signal should be altered,
our data augmentation technique applies the $f0$ and formant shifting
to the raw waveform, before extracting the features that are passed to the ASR model (see \S\ref{sec:models}).
The two operations are described below and have been implemented through Praat \cite{praat}.

\noindent\textbf{$f0$ Shifting.} 
First of all, we estimate the $f0$ values of the original audio, defining the $F\mathit{0}$ contour and its median $\tilde{f\mathit{0}}$. 
We then define a value representing the $f0$ median ($\tilde{f\mathit{0}}'$) of the new manipulated audio.
$\tilde{f\mathit{0}}'$ is sampled from a normal 
distribution whose mean and standard deviation depend on the target gender defined by the chosen policy. For female voices, we use 250 Hz as the mean and 17 as the standard deviation so that the sampled value is between 199 Hz and 301 Hz with 99.7\% probability. For male
voices, the mean is 140 Hz and the standard deviation is 20 to obtain a 99.7\% probability in the range [80, 200] Hz.
Once we have these values, we compute the scaling factor $\alpha$ as the ratio $\tilde{f\mathit{0}'} / \tilde{f\mathit{0}}$, 
by which we scale $F\mathit{0}$ to obtain the new target contour $F\mathit{0}'$.
Lastly, we alter the original waveform by integrating $F\mathit{0}'$ with the Time-Domain Pitch-Synchronous Overlap and Add (TD-PSOLA) synthesis \cite{psola}, 
as this approach 
manipulates the fundamental frequency,
without modifying the entire spectrum.

\noindent\textbf{Formant Shifting.} This effect is obtained by manipulating the sampling frequency. Specifically, the sampling frequency is scaled by a factor $\beta$, which determines a scaling of all the frequencies and the duration. 
$\beta$ is set to 1.2 if the target gender is female, 0.8 if it is male.
Lastly,
$f0$
and duration are scaled back to the original value using 
TD-PSOLA
(while formants remain manipulated), 
and the sound is 
restored
to the original sampling frequency via sinc interpolation \cite{shannon_sinc}.

\section{EXPERIMENTAL SETTINGS}
\label{sec:experimental-setup}

\subsection{Data and Evaluation Metrics}
\label{sec:data}

For training, we utilize the MuST-C corpus \cite{cattoni}, which is based on TED talks data and includes English audio, transcripts, and text translations into other languages. 
Specifically, we focus on the largest section, English-Spanish (en-es), and use its audio recordings (504h) along with their corresponding transcripts to train our ASR models.
The selection of MuST-C is driven by several factors.
First, 
it
provides reliable manual annotation of speaker gender, which 
eases the application of our data augmentation technique and the evaluation of the performance on each gender group data.
Second, it
exhibits a notable disparity between male and female segments, with approximately 70\% being male and 30\% female, which is representative of the currently-available data, making it an ideal testbed for evaluating our approach.\footnote{Librispeech \cite{Librispeech}, for instance, 
is designed to be
balanced in terms of female/male voices, hence representing an unrealistic and artificial setting.} 
Furthermore, MuST-C comprises spontaneous speech segments delivered by speakers in front of live audiences, and in public speech
there is typically a wide use of $f0$ variations and a broader range of $f0$ values, due to its richer repertoire of argumentative and emotionally colored elements \cite{public_speech}.
This characteristic helps to ensure that, also in a non-augmented condition, the models 
are trained on
various pitch variations.

We use the MuST-C dev set as validation set,
and we evaluate on the official tst-COMMON and tst-HE sets.
We used both test sets to maximize the pool of evaluation data available. 
Notably, tst-COMMON is larger (27 talks) but does not exhibit a balanced distribution in terms of speakers’ gender (30\% of segments uttered by female and 70\% by male speakers). 
In contrast, tst-HE (12 talks) offers a more 
balanced representation of
male and female speakers (45\% of segments are uttered by female speakers, and 55\% by male speakers), therefore being particularly suitable to study the effectiveness of our approach.
The evaluation 
involves
computing the word error rate (WER),\footnote{Computed with JiWER 2.3.0: https://github.com/jitsi/jiwer.} 
both on the overall test sets and separately for each gender group.
We also report the relative WER decrease (WERR) with respect to the baseline.

We encode text into BPE \cite{sennrich-etal-2016-neural} using SentencePiece \cite{kudo-richardson-2018-sentencepiece} with 8,000 vocabulary size, as per \cite{digangi}.
As audio features, we use
log-compressed mel-filterbanks (80 channels), computed
over windows of 25 ms with a stride of 10 ms using pykaldi.\footnote{https://github.com/pykaldi/pykaldi.}

\subsection{Models}
\label{sec:models}

Our architecture is made of two 1D convolutional layers with stride 2, which reduce the length of the input sequences by a factor of 4 as per \cite{fairseq-st}, 12 Conformer \cite{gulati} encoder layers, and 6 Transformer \cite{vaswani} decoder layers. 
The embeddings have 512 features and the inner feed-forward layers 2048. We set dropout to 0.1 and minimize label-smoothed cross entropy \cite{szegedy} with the Adam optimizer \cite{kingma} and Noam learning rate scheduler \cite{vaswani}, where the maximum learning rate is $2e^{-3}$.
The objective function also includes the sum with a Connectionist Temporal Classification (CTC) loss \cite{ctc}.
As default data augmentation strategy (even when we adopt our \textit{Opposite} and \textit{Random} policies),
we apply SpecAugment \cite{specaugment} to every input utterance, 
masking 1 segment over the frequency axis with maximum length 27,
and 1 segment over the time axis with maximum length 100. 
We train our models until the validation loss does not decrease for 10 
epochs,
and average the five checkpoints around the best on the validation set. 
The whole training takes $\sim$24 hours on 4 V100 GPU (16GB of VRAM).
Our code is based on fairseq-ST, 
and we use
the padding-safe Conformer implementation by \cite{papi2023reproducibility}.\footnote{Source code available at https://github.com/hlt-mt/FBK-fairseq/.}

\subsection{Hyperparameters}
\label{sec:hyperparam}

The two policies of our data augmentation technique, \textit{Opposite} and \textit{Random},
require the definition of respectively two ($p_{f\rightarrow{}m}$ and $p_{m\rightarrow{}f}$) and one ($p_r$) hyperparameters. To set their value for our experiments, we choose those with the lowest overall WER on the validation set.
\begin{table}[t]
\begin{minipage}[b]{0.45\linewidth}
\centering
\small
\begin{tabular}{cc|c}
$p_{f\rightarrow{}m}$ & $p_{m\rightarrow{}f}$ & \textbf{WER ($\downarrow$)}  \\
\hline
30\% & 70\% & \textbf{10.92} \\
70\% & 30\% & 11.08 \\
50\% & 50\% & 11.30 \\
70\% & 70\% & 11.36 \\
100\% & 100\% & 11.91
\end{tabular}
\caption{\label{tab:opposite_dev} 
WER on the dev set for \textit{Opposite} with different ($p_{f\rightarrow{}m}$, $p_{m\rightarrow{}f}$) pairs.}
\end{minipage}%
\hfill
\begin{minipage}[b]{0.45\linewidth}
\centering
\small
\begin{tabular}{c|c}
$p_r$ & \textbf{WER ($\downarrow$)} \\
\hline
30\% & 11.21 \\
40\% & 11.16 \\
50\% & \textbf{10.89} \\
60\% & 11.12 \\
70\% & 11.05 \\
80\% & 11.02 \\
100\% & 11.21 \\
\end{tabular}
\caption{\label{tab:random_dev}
WER on the dev set for \textit{Random} with different $p_r$ values.}
\end{minipage}
\end{table}
For the \textit{Opposite} policy, we test the following values for 
$(p_{f\rightarrow{}m}, p_{m\rightarrow{}f})$: (30\%, 70\%), (70\%, 30\%), (50\%, 50\%), (70\%, 70\%), (100\%, 100\%).
As we can see in Table \ref{tab:opposite_dev}, the best WER is obtained with (30\%, 70\%).
As 70\% of MuST-C is made of male
voices, with this combination the amount of female/male voices in the training data is perfectly balanced.
For the \textit{Random} policy, we test all values between 30\% and 80\% with a 10\% interval for $p_r$. 
Results in Table \ref{tab:random_dev} show that we obtain the lowest WER with $p_r$ at 50\%. 
This means that the training data still contains more male
samples (parity is reached only at 100\%), but the unbalance is halved with respect to the original training data, with a distribution of around 40\%-60\% in favor of male speakers.
Hence, we provide results and analyses based on \textit{Opposite} policy with $(p_{f\rightarrow{}m}, p_{m\rightarrow{}f}) =$ (30\%, 70\%), and \textit{Random} policy with $p_r =$ 50\%.

\begin{table*}[t]
\small
\centering
\begin{tabular}{l|ccc|ccc||ccc|ccc}
 & \multicolumn{6}{c||}{\textbf{tst-COMMON}}   & \multicolumn{6}{c}{\textbf{tst-HE}}  \\
 & \multicolumn{3}{c|}{\textbf{WER ($\downarrow$)}}   & \multicolumn{3}{c||}{\textbf{WERR\% ($\uparrow$)}}  & \multicolumn{3}{c|}{\textbf{WER ($\downarrow$)}}   & \multicolumn{3}{c}{\textbf{WERR\% ($\uparrow$)}}  \\
 & Overall & F  & M  & Overall   & F   & M  & Overall & F  & M  & Overall   & F   & M\\
\hline
Baseline  & 10.10  & 8.51  & 10.79 & 0.00 & 0.00 & 0.00 & 7.53 & 8.72 & \textbf{6.59} & 0.00 & 0.00 & \textbf{0.00} \\
 + VTLP  & 10.30  & 8.61  & 11.03 & -1.98 & -1.18 & -2.22 & 8.24 & 9.04 & 7.61 & -9.43 & -3.67 & -15.48 \\

\hline
Random & \textbf{9.65*} & \textbf{7.67}* & \textbf{10.50}* & \textbf{4.46} & \textbf{9.87} & \textbf{2.69} & \textbf{7.28}  & \textbf{8.10}* & 6.62 & \textbf{3.32} & \textbf{7.11} & -0.04 \\
Opposite & 9.72* & 7.70*  & 10.59 & 3.76 & 9.52 & 1.85 & 7.63 & 8.40 & 7.02* & -1.32 & 3.67 & -6.52 \\
\hline
\end{tabular}
\caption{\label{tab:results} WER and WERR on the MuST-C tst-COMMON and tst-HE sets of the en-es section with (\textit{Random} and \textit{Opposite}) and without (\textit{Baseline} and \textit{Baseline + VTLP}) our data augmentation method. Results are reported also on the female (F) and male (M) portions. 
*~indicates that the improvement
 or the degradation 
of our augmentation techniques over \textit{Baseline}} is statistically significant (bootstrap resampling \cite{koehn-2004-statistical} with 95\% CI and 1,000 samples).
\end{table*}

\begin{table*}[t]
\small
\centering
\begin{tabular}{l|ccc|ccc||ccc|ccc}
 & \multicolumn{6}{c||}{\textbf{tst-COMMON}}   & \multicolumn{6}{c}{\textbf{tst-HE}}  \\
 & \multicolumn{3}{c|}{\textbf{WER ($\downarrow$)}}   & \multicolumn{3}{c||}{\textbf{WERR\% ($\uparrow$)}}  & \multicolumn{3}{c|}{\textbf{WER ($\downarrow$)}}   & \multicolumn{3}{c}{\textbf{WERR\% ($\uparrow$)}}  \\
 & Overall & F  & M  & Overall   & F   & M  & Overall & F  & M  & Overall   & F   & M\\
\hline
\hline
Random & \textbf{9.65} & \textbf{7.67} & \textbf{10.50} & \textbf{4.46} & \textbf{9.87} & \textbf{2.69}  & \textbf{7.28}  & \textbf{8.10} & \textbf{6.62} & \textbf{3.32} & \textbf{7.11} & \textbf{-0.04}  \\
 - Formant Shifting & 9.80 & 7.98 & 10.58 & 2.97 & 6.23  & 1.95 & 7.37 & 8.33 & \textbf{6.62} & 2.13 & 4.47 & \textbf{-0.04} \\
\hspace{2mm}  - Gender Switching
& 10.14  & 8.46 &  10.86 & -0.40  & 0.59 & -0.65  & 7.57 & 8.33 & 6.97 & -0.53 &  4.47 & -5.77 \\
 \hline
\end{tabular}
\caption{\label{tab:ablation} Ablation study for the \textit{Random} policy with WER and WERR (compared to \textit{Baseline}) on the en-es MuST-C tst-COMMON and tst-HE sets. 
Results are also reported on the female (F) and male (M) portions.}
\end{table*}

\section{RESULTS}
\label{sec:results}

\subsection{Main Results}

In Table \ref{tab:results}, we report the WER and WERR 
results computed on the MuST-C tst-COMMON and tst-HE sets,
both for the \textit{Random} and the \textit{Opposite} policies.
We compare them to two baselines: 
one (\textit{Baseline}) without any pitch perturbation strategy;
the other (\textit{Baseline + VTLP}) with VTLP.\footnote{
Following \cite{jaitly}, we warp each utterance with a factor sampled from a normal distribution (mean of 1 and standard deviation of 0.1), but constrained within the range [0.9, 1.1], and a boundary frequency of 4800 Hz.}
We implemented the latter to assess whether our gender-specific approach outperforms generic pitch perturbation methods.

Firstly, it is worth noting that our \textit{Baseline} outperforms by a margin of 0.42 the best published result (to the best of our knowledge) on tst-COMMON \cite{papi2023reproducibility}, which achieved a WER score of 10.52.
This demonstrates the validity of our experimental settings.
Secondly, VTLP does not provide any 
benefit
and, in fact, \textit{Baseline + VTLP} increases the WER for both test sets and across genders.
Regarding the two gender subgroups, all systems have higher WER for female speakers than for male speakers in tst-HE, as expected, whereas the trend is the opposite in tst-COMMON. We speculate that this difference 
may be due to factors like speakers' accents.
In tst-HE, most (10) of the 12 speakers are indeed American, 
one is British (in the female subset) and one is Pakistani (in the male subset).
In contrast, tst-COMMON features a broader range of accents, with only one-third of the speakers being American in both gender groups. 
The male subgroup includes speakers with diverse accents, from 
Northern
Europe to Kenya, while three out of the five non-American female speakers are native English speakers (Canadian and Australian).
This accent variability can significantly affect performance.

Looking at the comparison between
\textit{Baseline}
and our solution, we notice that both our methods (\textit{Random} and \textit{Opposite})
always improve the recognition accuracy in the female subsets. 
In
the male subsets, instead,
\textit{Baseline}
is outperformed
on tst-COMMON, while on tst-HE
it is on par with \textit{Random}.
The different behavior between the two sets on male speakers may be motivated by the wider range of variations in the tst-COMMON set,
where non-native speakers are highly represented.
Interestingly,
even in tst-COMMON, where the WER on female samples is lower than that of male samples,
the gains are always 
higher on
the female portion.

Among the policies, \textit{Random} proves to be consistently the best on all test sets and genders. 
Its superior scores demonstrate that,
although \textit{Opposite} produces a training set with a more balanced 50\%-50\% distribution compared to 
\textit{Random}'s 40\%-60\% distribution,
the higher variability introduced by \textit{Random}, which also alters samples within the same gender group, proves to be more effective for both female and male speech.
The WER differences between \textit{Random} and \textit{Baseline} are marginal in the male portions,
showing limited yet statistically significant gains on tst-COMMON (2.69\% WERR), and comparable results on tst-HE (-0.04\% WERR).
Conversely, the improvements in the female segments are statistically significant, with error reductions
ranging from 7.11\% to 9.87\%. 
These gains translate into an overall WERR of 4.46\%-3.32\%.

All in all, we can conclude that our data augmentation technique with the \textit{Random} policy improves the recognition accuracy of female voices without negatively affecting the recognition of male voices (where it also provides marginal gains).
This not only produces benefits to the overall performance but leads to fairer and more robust systems, even when data for one gender is scarce.

\subsection{Ablation Study}

We conduct an ablation study on our best policy (\textit{Random}) to better understand how each design choice impacts the results.
The scores are reported
in Table \ref{tab:ablation}.
In particular, we first remove the formant shifting operation (\textit{- Formant Shifting}), to assess whether the
$f0$ shifting alone is enough. Second, in addition to shifting only the 
$f0$,
we always sample the new 
$f0$
median frequency
from
the range corresponding to the same gender 
(\textit{- Gender Switching})
instead of randomly picking it from either the female or male frequency range.

Looking at the overall WER, we notice that both formant shifting and gender switching contribute to increasing recognition accuracy. 
Specifically, formant shifting is critical mostly in the female subset, where it accounts for
$\sim$37\%
of the relative improvement with respect to the baseline (WERR 
decreases
from 9.87 to 6.23 on tst-COMMON and from 7.11 to 4.47 on tst-HE). 
The male subset, instead, seems not to benefit from formant shifting: on tst-HE the WER is identical, and on tst-COMMON the gap is minimal. 

When we also force choosing 
an $f0$
frequency typical of the same gender (\textit{- Gender Switching}), the overall WER gains vanish, demonstrating that small
$f0$
variations are not enough to improve the performance of ASR systems.
For female voices, slight improvements
are
still noticeable compared to 
\textit{Baseline}
(0.59\% for tst-COMMON and 4.47\% for tst-HE).
However, the large drops with respect to \textit{- Formant Shifting}
attest that
rebalancing the training data distribution by shifting across genders is essential to yield considerable improvements.
For male voices, instead, scores are consistently
lower also compared to
\textit{Baseline}
(negative WERR).
Thus, $f0$ shifting within the same gender group proves to be counter-productive for the most represented category and at least $f0$ shifting across genders is required to retain the performance.

Overall, the ablation study emphasizes the 
need for perturbing both $f0$ and formants,
as well as the importance of shifting voices from one gender to the other. 
Our approach would not yield improvements without these design choices.

\pgfplotstableread[row sep=\\,col sep=&]{
Bins & WERRM & WERRF & WordCountM & WordCountF \\
80                      & 0                           &                             & 13                           &                              \\
90                      & -10.52                      &                             & 97                           &                              \\
100                     & 7.31                        &                             & 746                          &                              \\
110                     & 10.35                       &                             & 2299                         & 27                           \\
120                     & 1.45                        &                             & 3408                         & 208                          \\
130                     & -0.78                       &                             & 6475                         & 426                          \\
140                     & 2.03                        &                             & 7373                         & 434                          \\
150                     & 4.61                        &                             & 7554                         & 362                          \\
160                     & 2.60                        &                             & 5200                         & 350                          \\
170                     & -11.41                      & 8.28                        & 3161                         & 793                          \\
180                     & 13.29                       & 0                           & 1555                         & 726                          \\
190                     & 4.06                        & 3.24                        & 382                          & 1155                         \\
200                     & 8.28                        & 18.66                       & 305                          & 1800                         \\
210                     & 6.83                        & 19.03                       & 157                          & 1437                         \\
220                     & 11.80                       & 13.01                       & 118                          & 1320                         \\
230                     &                             & 12.12                       & 40                           & 1710                         \\
240                     &                             & 2.75                        & 42                           & 2392                         \\
250                     &                             & 6.20                        & 13                           & 2638                         \\
260                     &                             & 7.81                        &                              & 1944                         \\
270                     &                             & 20.19                       &                              & 825                          \\
280                     &                             & 14.59                       &                              & 555                          \\
290                     &                             & -9.49                       &                              & 127                          \\
300                     &                             & 6.24                        &                              & 74                           \\
310                     &                             &                             &                              &                              \\
320                     &                             & 0                           &                              & 11     \\
}\tstCOMMONwerr

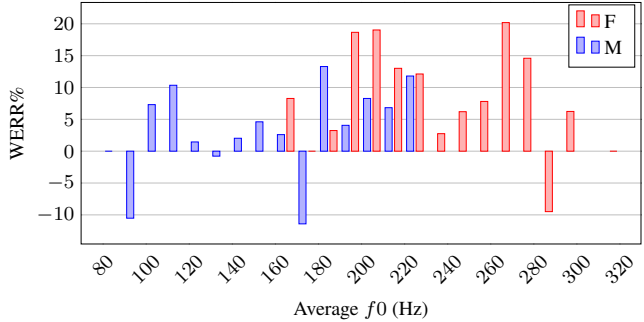
\begin{figure}[t]
\centering
    \footnotesize
\begin{tikzpicture}[scale=0.94]
\begin{axis}[
            ybar,
            bar width=.1cm,
            legend style={
                at={(0.93,0.99)}, 
                anchor=north, font=\footnotesize},
            legend cell align={left},
            tickwidth         = 0pt,
            ytick={-10, -5, 0, 5, 10, 15, 20},
            ylabel={WERR\%},
            ylabel near ticks,
            xlabel={Average $f0$ (Hz)},
            xlabel near ticks,
            xtick distance=20,
            xmin=70,
            xmax=330,
            style={outer sep=0},
            height=5cm,
            width=9.5cm,
            xticklabel style={rotate=45},
            ymajorgrids=true,
        ]
        \addplot[color=red, fill=red!30!white] table[x=Bins,y=WERRF]{\tstCOMMONwerr};
        \addplot[color=blue, fill=blue!30!white] table[x=Bins,y=WERRM]{\tstCOMMONwerr};
        \legend{F, M}
    \end{axis}
\end{tikzpicture}
\caption{\label{fig:tstCOMMON_analysis} 
WERR of \textit{Random} compared to \textit{Baseline} on tst-COMMON and tst-HE (combined) for female (F) and male (M) speakers across various average $f0$ ranges.}
\end{figure}

\begin{figure}[t]
\centering
    \footnotesize
\begin{tikzpicture}[scale=0.94]
\begin{axis}[
            legend style={
                at={(0.91,0.99)}, 
                anchor=north, font=\footnotesize},
            legend cell align={left},
            tickwidth         = 0pt,
            ylabel={Num. words},
            ylabel near ticks,
            xlabel={Average $f0$ (Hz)},
            xlabel near ticks,
            xtick distance=20,
            xmin=70,
            xmax=330,
            style={outer sep=0},
            height=5cm,
            width=9.3cm,
            xticklabel style={rotate=45},
            /pgf/number format/1000 sep={},
            ymajorgrids=true,
        ]
        \addplot[mark=none, color=red] table[x=Bins,y=WordCountF]{\tstCOMMONwerr};
        \addplot[mark=none, color=blue] table[x=Bins,y=WordCountM]{\tstCOMMONwerr};
        \legend{F, M}
    \end{axis}
\end{tikzpicture}
\caption{\label{fig:tstCOMMON_wc} Distribution of the number of words across the average $f0$ ranges for tst-COMMON and tst-HE (combined), divided between female (F) and male (M)  speakers.}
\end{figure}
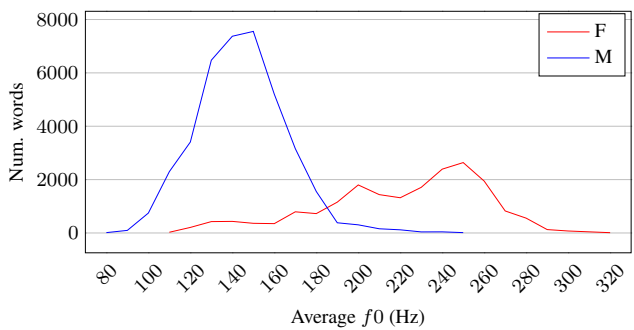

\section{IMPACT ACROSS FREQUENCY RANGES}
\label{sec:analysis}

To better understand the source of recognition gains of our \textit{Random} policy,
we also analyse performance variations
as the average $f0$ varies across segments.
Figure \ref{fig:tstCOMMON_analysis} illustrates
the WERR values of \textit{Random} compared to \textit{Baseline} for the segments of tst-COMMON and tst-HE (combined), grouped by average $f0$ ranges.
As a reference, in Figure \ref{fig:tstCOMMON_wc} we also
display the word counts per segments grouped by average $f0$ in the two combined test sets.

Positive WERRs are observed across all ranges, with only three isolated negative values, 
specifically around 90 Hz and 170 Hz for male voices, and 290 Hz for female voices, which are likely outliers caused by the limited amount of test data in those ranges 
(see Figure \ref{fig:tstCOMMON_wc}).
However, significant improvements (around or higher than 10\%) were observed in specific ranges, such as 100-110 Hz, 180 Hz and 220 Hz for male voices, and 200-230 Hz and 270-280 Hz for female voices.
Interestingly, these notable improvements do not align with the median of the $f0$ distributions for men and women.
Instead, they correspond to the boundary of the male and female $f0$ ranges, describing an antiphase-like pattern with respect to the distribution curves.
In particular, the male area spanning 200-220 Hz and the female area spanning 270-280 Hz represent less typical frequencies. 
Additionally, the 200-220 Hz range is a span where the two gender $f0$ distributions overlap, showing considerable and consistent gains on both genders.

Overall, although improvements are observed across all ranges, some notable peaks are located
at the boundaries of the typical $f0$ values for male and female voices. 
As these frequency spans are less common, they are less represented in the data.
The random perturbations introduced by our data augmentation strategy increase the coverage of these less-likely ranges and of
a more diverse set of pitch variations. 
This results in enhanced recognition accuracy,
especially in regions where, regardless of gender, data might be limited.

\section{CONCLUSION}
\label{sec:conclusion}

We presented a data augmentation technique designed to enhance the robustness of ASR models to female voices, which are often under-represented in speech training data. Our method involves manipulating critical acoustic parameters, namely the fundamental frequency and formants, which distinguish male and female speech. By introducing perturbations to these parameters, we generate voice variants for both genders, using samples from both the same and opposite genders. This approach increases the variability within each gender group and helps balancing 
the gender distribution in the training data.
Experimental results on English public speech demonstrate 
that our technique significantly improves
the recognition accuracy of female voices, with relative WER improvements up to 9.87\%. 
The improvement does not come at the detriment of performance on male voices, which indeed features limited benefits. Additionally, when analyzing the WERR values across different $f0$ ranges, we observed larger gains in regions considered less typical or potentially ambiguous for both genders. 
As our approach effectively addresses gender bias toward fairer ASR systems, our future plan is to extend it to closely related tasks, like speech translation, where the problem is also attested and pervasive \cite{mustshe}.

\section{ACKNOWLEDGEMENTS}
\label{sec:acknoledgements}
This work is part of the project “Bias Mitigation and Gender Neutralization Techniques for Automatic Translation”, which is financially supported by an Amazon Research Award AWS AI grant. 
We also acknowledge the support of the PNRR project FAIR - Future AI Research (PE00000013), under the NRRP MUR program funded by the NextGenerationEU.

\bibliographystyle{IEEEbib}
\bibliography{custom}

\end{document}